\pdfoutput=1

\documentclass[11pt]{article}

\usepackage[]{emnlp2021}

\usepackage{times}
\usepackage{latexsym}
\usepackage{booktabs}
\usepackage{amsmath}
\usepackage{amssymb}
\usepackage{graphicx}
\usepackage{multirow}
\usepackage{multicol}
\usepackage{subcaption}
\usepackage{caption}
\usepackage{hyperref}
\usepackage{cleveref}

\usepackage[T1]{fontenc}

\usepackage[utf8]{inputenc}

\usepackage{microtype}

\usepackage{times}
\usepackage{graphicx}
\usepackage{latexsym}
\usepackage{natbib}
\usepackage{amsmath}
\usepackage{amssymb}
\usepackage{mathtools}
\usepackage{anyfontsize}
\usepackage{booktabs}
\usepackage{csquotes}
\usepackage{multirow}
\usepackage{array}
\usepackage{parskip}
\usepackage{setspace}
\usepackage{graphicx}
\usepackage{subcaption}
\usepackage{color,soul}

\usepackage{microtype}

\usepackage{amsmath,amsfonts,bm}

\def\eqref#1{equation~\ref{#1}}

\def\1{\bm{1}}

\def\mE{{\bm{E}}}

\DeclareMathAlphabet{\mathsfit}{\encodingdefault}{\sfdefault}{m}{sl}
\SetMathAlphabet{\mathsfit}{bold}{\encodingdefault}{\sfdefault}{bx}{n}

\def\sR{{\mathbb{R}}}

\newcommand{\insertappendix}{

\appendix

\section{Designing the Subformer}

\paragraph{Naming} The Subformer is a play on words, referencing its small size - i.e. \textit{sub-}, as well as the \textit{Sandwich}-style parameter sharing technique, referencing the type of sandwich.

\paragraph{Architecture}

When using SAFE, our parameter count would result in $V \times d_e + \ 5d_e^2 + d_e \times d_m$ parameters, where the first term represents the embedding layer, the value $5d_e^2$ groups the query, key, and value projections and 2 output feed-forward layers, and $d_e \times d_m$ represents the linear projection from the embedding dimension to the model dimension. As mentioned in the paper, this results in a significant parameter reduction for values of $d_e \ll d_m$. 

As we tie the decoder's output projection layer (returning a distribution over the vocabulary) with the decoder's input embedding matrix, we project the decoder's last hidden state (with dimension $d_m$) to $d_e$ using a two layer MLP. Also, when we perform encoder attention in the decoder's Sandwich Module, we simply linearly project the query from the decoder from $d_s$ to $d_m$ and then project it back to $d_s$ once the attention operation is complete.

\paragraph{Memory Footprint}

\begin{table*} \footnotesize \centering
    \begin{tabular}{lcc}
        \toprule [0.2em]
	    \textbf{Method} & \textbf{Embedding Memory Usage} & \textbf{Model Memory Usage} \\
        \midrule[0.1em]
	    Transformer & $d_m \times V$ & $L(4d_m^2 + 2\big(\vec{d}_m \times d_m)\big)$ \vspace{1mm}\\
	    Sandwich (naive) & --- &  $3(4d_m^2 + 2(4\vec{d}_m \times d_m)\big) \vspace{1mm}$ \\
	    Subformer & $d_e  \times V+ 5d_e^2 + d_e \times d_m$ &  {\footnotesize$2\big(4d_m^2 + 2(\vec{d}_m \times d_m)\big) + \big(4d_s^2 + 2\big(\vec{d}_s \times d_s + 2(d_s\times d_m)\big)\big)$} \vspace{1mm}\\
        \bottomrule[0.2em]
    \end{tabular}
    \caption{Memory space required by each method given a stack of encoder layers. \textit{Sandwich (naive)} refers to simply performing Sandwich style parameter sharing with no other modifications to the architecture.} \label{tb:mem}
\end{table*}

Table~\ref{tb:mem} summarizes the memory footprint of our proposed techniques. In this table, the benefits of Sandwich-style parameter sharing can be seen as the number of independent layers is controlled to be $L \leq 3$, however, Transformers generally need to be deeper (with a standard of $L = 6$) to learn more meaningful representations with the parameter count scaling linearly with respect to the layer count. Similarly, the benefits of disentangling the model dimension from the embedding dimension can be seen as well. Due to the parameter reduction attained by these techniques, the models can be trained in memory-constrained scenarios with a larger batch size.

\section{Data and Training Details}\label{app:training_details}

Training was done on 8 GPUs on a single DGX-1 Machine, while training on 16 GPUs was done using multiple compute nodes of a compute cluster. We train all base/small models on 8 NVIDIA Tesla V100 GPUs. For all big/large models, we train on 16 NVIDIA Tesla V100 GPUs. All models were trained with mixed precision \citep{micikevicius2017mixed} and are implemented in PyTorch \citep{paszke_pytorch_2019} using our modification of \texttt{fairseq} \citep{ott2019fairseq}.

\paragraph{Machine Translation} 

We train using 8192 tokens per GPU an update frequency of 2, for small, base models. For large models, we train on 16 GPUs with 4096 tokens per GPU with an update frequency of 2. We follow the training setup of \citet{ghazvininejad2019mask}: we use the same weight initialization scheme as BERT \citep{devlin2019bert}, sampling weights from $\mathcal{N}(0,0.02)$, initializing biases to zero and setting layer normalization parameters $\beta$ and $\gamma$ to be $0$ and $1$, respectively. For regularization we use the best of $[0.1,0.2,0.3]$ dropout, weight decay of $0.01$, while using label-smoothed cross entropy loss with $\epsilon = 0.1$. We train using an effective batch size of 128K tokens. The models are trained using Adam \citep{kingma2014adam}, with hyper-parameters $\beta = (0.9,0.999)$ and $\epsilon = 10^{-6}$. We warm up the learning rate to a peak of $5 \times 10^{-4}$ within 10K iterations and then decay the learning rate with the inverse square root schedule. When creating the final model, we use the checkpoint with the lowest loss on the development set and generate using a beam size of 5 \citep{vaswani2017attention}, tuning the length penalty of $\alpha \in [0.0,0.2,$\dots$,2.0]$ in the validation set. We perform early stopping, training for a maximum of 250K iterations.

We use the following settings for our models: (1) Subformer-small has $d_m = 512$, $d_s = 768$, $d_{e} = 256$ and $L=8$, (2) Subformer-base has $d_m = 512$, $d_s = 1024$, $\vec{d}_s = 3072$, $d_e = 320$, (3) Subformer-mid has $d_m = 768$, $d_s = 768$, $d_e = 350$ and (4) Subformer-large has $d_m = 1024$, $d_s = 2048$ and $d_e = 512$. For WMT'16 EN-RO, our small model has $d_m = 320$, $d_s = 512$ and $d_{e} = 192$ and our base model has $d_m = 512$, $d_s = 640$, and $d_e = 384$.

In terms of datasets, we make use of the same pre-processed data used by \citet{ghazvininejad2019mask} for WMT'14 EN-DE with a 32K BPE \citep{sennrich2016neural} vocabulary and during evaluation we perform de-hyphenation \citep{vaswani2017attention}. We use the same data as \citet{lee2018deterministic} for WMT'16 EN-RO with a 35K BPE vocabulary.

\paragraph{Abstractive Summarization}
We follow \citet{edunov2019pre} and use the official \texttt{ROUGE-1.5.5.pl} script with parameters \texttt{-m -a -n 2}. As mentioned in the paper, our model configuration is the same as Subformer-base, but we set $d_e = 256$. Articles are truncated to 400 tokens \citep{see2017get} and we use a BPE vocabulary of 32K types \citep{edunov2019pre}. We follow the training schedule of \citet{edunov2019pre}. During inference, we tune generation length in the range of \{40, 50, 60\} and use tri-gram blocking, following standard practice. When pretraining, we pretrain on Wikipedia (14GB) for 100K iterations, using a batch size of 512K tokens. We use a learning rate of 7e-4, warmed up over 10K iterations. %
 
\paragraph{Language Modeling}
When training our language models, we use 8 GPUs with 3072 tokens per GPU and an update frequency of 3, following \citet{baevski2018adaptive}. Models are trained using Nesterov's accelerated gradient optimizer \citep{sutskever2013importance}, warming up the learning rate to 1.0 for 16K iterations, and then annealing for 270K iterations using a cosine annealing schedule. We use three configurations: (1) $d_m = 768, \vec{d}_m = 3072, d_s = 1536, \vec{d}_s = 6144$, (2) $d_m = 768, \vec{d}_m = 4096$ and $d_s = 2048, \vec{d}_s = 6144$ and (3) $d_m = 1024, \vec{d}_m = 4096$ and $d_s = 2048, \vec{d}_s = 6144$. All models use $L = 12$. Our considered dataset, Wikitext-103, contains 103M tokens and has a vocabulary of nearly 270K.

\section{Extended Related Work}

\paragraph{Improving Transformers}
Given the effectiveness of the Transformer, improving the architecture has been of much interest to the NLP community. Within this domain, one branch of research concerns the reduction of the quadratic complexity (w.r.t. sequence length) of the Transformer's core self-attention mechanism \citep{wu2019pay,kitaev2020reformer}, pushing it down to linear or log-linear complexity. The second branch of research regards improving the expressiveness of Transformer models, by using more layers \citep{Dou_2018}, or by improving the architecture \citep{wu2019pay,so2019evolved}. The third branch of research regards improving the parameter efficiency of Transformers. Approaches towards this goal include neural architecture search approaches \citep{so2019evolved,wu2020lite}, where new Transformer-based architectures are learned using gradient descent, more manually crafted approaches \citep{dehghani2018universal,mehta2020delight}, as well as weight-sharing approaches \citep{lan2020albert,wu2019pay}. The work most similar to ours is ALBERT \citep{lan2020albert} in which complete weight sharing is used to pre-train deep contextualized word representations \citep{Peters:2018,devlin2019bert}. Different from this work, we focus on common NLP generative/sequence-to-sequence tasks versus large-scale pre-training and develop an approach to increase model capacity while reducing parameter footprint tailored to this setting.

\paragraph{Compressing Transformers} We find prior work on pruning and quantizing Transformer models to reduce their size with a focus on sequence-to-sequence settings like machine translation \citep{pratoFullyQuantizedTransformer2019}, on encoder-based methods like BERT \citep{zafrirQ8BERTQuantized8Bit2019,ganeshCompressingLargeScaleTransformerBased2020} or with a more generic scope in mind \citep{cheong2019transformers,leeSNIPSingleshotNetwork2018}. Our approach is orthogonal to these since we directly aim at reducing the number of parameters of Transformer models by proposing architecture modifications and weight sharing techniques.

\paragraph{Reducing Embedding Dimensionality in Sequence Models}
As embeddings can substantially increase the parameter count as the vocabulary size increases, especially in sequence modeling scenarios, embedding reduction techniques have been proposed, including using a linear projection to project to a lower dimension \citep{baevski2018adaptive,dai2019transformer} or using combinations of block sparse transformations \citep{mehta2020define,mehta2020delight}. We propose a self-attention based projection layer, SAFE, which we empirically show to outperform the aforementioned linear projection methods with a similar parameter count. 

}

\title{Subformer: Exploring Weight Sharing for Parameter Efficiency in Generative Transformers}

\author{Machel Reid$^\vartheta$, Edison Marrese-Taylor$^{\aleph,\vartheta}$, Yutaka Matsuo$^\vartheta$\\
$^\vartheta$The University of Tokyo\\
$^\aleph$National Institute of Advanced Industrial Science and Technology (AIST) \\
{\tt\{machelreid,emarrese,matsuo\}@weblab.t.u-tokyo.ac.jp}}

\begin{document}
\maketitle
\begin{abstract}
Transformers have shown improved performance when compared to previous architectures for sequence processing such as RNNs. Despite their sizeable performance gains, as recently suggested, the model is computationally expensive to train and with a high parameter budget. In light of this, we explore parameter-sharing methods in Transformers with a specific focus on generative models. We perform an analysis of different parameter sharing/reduction methods and develop the Subformer. Our model combines sandwich-style parameter sharing, which overcomes naive cross-layer parameter sharing in generative models, and self-attentive embedding factorization (SAFE). Experiments on machine translation, abstractive summarization and language modeling show that the Subformer can outperform the Transformer even when using significantly fewer parameters.\footnote{\url{https://github.com/machelreid/subformer}}
\end{abstract}

\section{Introduction}

Recent improvements in NLP tasks can be attributed to the Transformer \citep{vaswani2017attention} model. The model has led to better deeply contextualized representations \citep{devlin2019bert}, better machine translation systems \citep{vaswani2017attention}, and language models \citep{baevski2018adaptive,dai2019transformer}. Despite their success, one main drawback of training these models is their computational cost, being a greatly limiting factor for many, with training times and memory usage ballooning as model sizes increase to attain better performance. 

With this in mind, there has been recent interest in making the Transformer more parameter-efficient to reap its performance benefits while making the model more computationally efficient and able to scale better. Many approaches have focused on automating model design with neural architecture search that aim at finding more efficient Transformer variations using gradient descent \citep{wu2020lite,so2019evolved,mehta2020delight}. As such, these techniques are expensive, requiring a significant amount of GPU hours to find good designs. In contrast to these approaches, the model by \citep{lan2020albert} has directly tackled model parameter reduction in the context of deeply contextualized word representations, still attaining similar (or better) performance. In this paper, we take a similar approach and look to explore whether these ideas can be applied to sequence-to-sequence models in a simple manner by designing the Subformer.%

The Subformer incorporates two novel techniques: (1) SAFE (Self-Attentive Factorized Embeddings), in which we use a small self-attention layer to reduce embedding parameter count, and (2) Sandwich-style Parameter Sharing, in which we develop a simple and intuitive technique for parameter sharing to be effective in Transformer models.

We evaluate the Subformer on machine translation, abstractive summarization, and language modeling, showing that our model can achieve similar or better performance compared with a base/big Transformer with a $\sim$40\% parameter reduction and minimal modification to the original architecture, reinforcing the existing over-parameterization claims \citep{fan2020reducing,mehta2020delight,lan2020albert}. On WMT'14 EN-DE we achieve a BLEU score of 29.3, compared to Transformer-big's 28.6 with 13M fewer parameters, and we also outperform the Transformer-XL model with a significant 3.6 lower perplexity and 37\% fewer parameters.

\section{The Subformer}
Let us start by defining the notation to be used throughout the paper. We refer to the model dimension as $d_m$, feed-forward projection dimension as $\vec{d}_m$, the vocabulary size as $V$, and the number of layers as $L$. Note that, unlike standard Transformer models, in which the embedding dimension is kept the same as $d_m$, we disentangle to embedding dimension to reduce parameter count. For this reason, we denote the embedding dimension to be $d_e$. 

\subsection{SAFE: \textbf{S}elf-\textbf{A}ttentive \textbf{F}actorized \textbf{E}mbeddings} \label{sec:safe}
We propose to reduce the number of parameters in our embedding layers, which can take up to 25\%  of the total parameter count (in the case of Transformer-base, \citealp{vaswani2017attention}), using a small self-attention layer. Specifically, we look to reduce the embedding size by disentangling the model dimension from the embedding dimension, reducing the embedding dimension $d_e$, and then projecting this to the model dimension $d_m$ using a small self-attention sub-layer followed by a feed-forward module. 

Given vocabulary size $V$, the usage of a standard embedding layer would result in $V \times d_m$ parameters. However, considering that the power of Transformers lies in their ability to learn contextual representations with attention, using a smaller $d_e$ for non-contextual embeddings and then projecting to $d_m$ is intuitively an effective method for parameter reduction \citep{lan2020albert}. This results in a significant parameter reduction for values of $d_e \ll d_m$.

\begin{figure}[h!]
    \centering
    \includegraphics[width=0.48\textwidth]{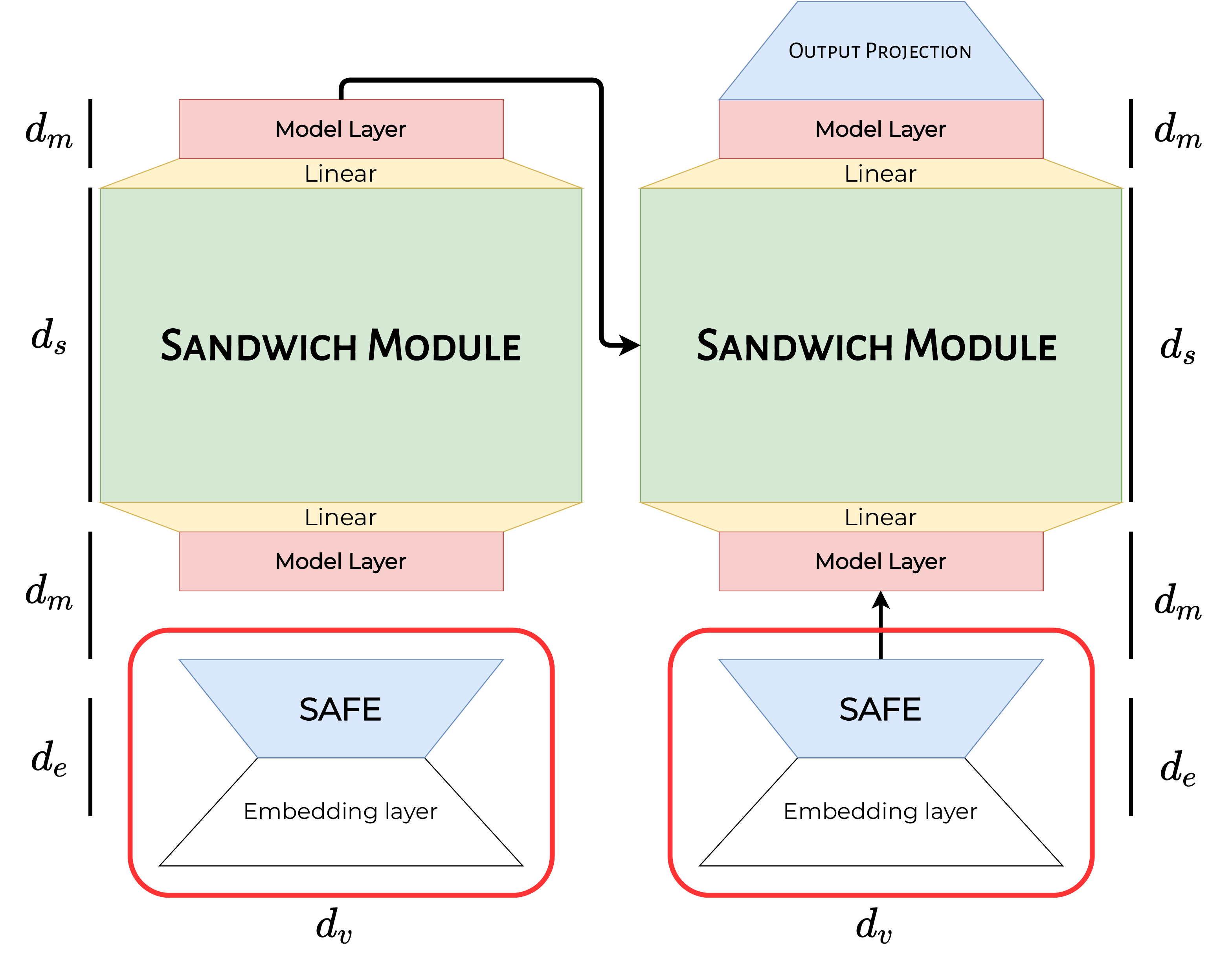}
    
    \caption{\textbf{The Subformer} with its four main components: (1) \textbf{\textit{SAFE embeddings}} (colored blue) and output projection layers, (2) \textbf{\textit{model layers}} which are placed at the top and bottom of the model (colored red), (3) \textbf{\textit{Sandwich Module}}, in which a wider shared layer composes the central part of our encoder/decoder, (4) \textbf{\textit{ projection layers}}, which allow for the interaction between the model layers and Sandwich Module despite their different dimensions (colored yellow).}
    \label{fig:diagram}
\end{figure}

When using SAFE, the embedding layer is composed of a regular token $\rightarrow$ vector embedding matrix $\mE \in \sR^{V \times d_e}$. This is followed by projecting the embeddings (summed with the positional encodings \citep{vaswani2017attention}, denoted by $PE$) to the model dimension $d_m$ using SAFE. Once we have our SAFE embeddings, we feed them through the first model layer ---the base of the sandwich. The output of this first layer is then projected to the sandwich dimension $d_s$. Once fed through the shared sandwich layers, we then project the output back to the model dimension using an MLP. The output of the projection is then fed through the final model layer to produce the output vectors.

Current models \citep{baevski2018adaptive,dai2019transformer,lan2020albert} often use a single linear projection, i.e. $V \times d_e + d_e \times d_m$. In contrast, we empirically show that simply contextualizing this projection with a small self-attention layer results in stronger performance with a minimal addition of parameters ---especially in the encoder-decoder case, where the input embedding layer and output projection are often tied \citep{press-wolf-2017-using} (Table~\ref{tb:safe}).
	
\subsection{Sandwich-style Parameter Sharing}%
Weight sharing techniques, despite being surprisingly effective, have been relatively unexplored in the context of generative Transformers. However, this has been shown to be powerful for leveraging models with large capacity and less memory usage/computation \citep{wu2019pay,lan2020albert}. 

\begin{table}[t]
    \centering
    \small
    \begin{tabular}{lcc}
        \toprule [0.2em]
        \multirow{1}{*}{\textbf{\textsc{Model}}} %
    	& \textbf{Param.} & \textbf{BLEU} \\
    	\midrule[0.1em]
    	DeFINE \citep{mehta2020define} & 52M & 27.0 \\
    	\midrule[0.07em]
    	$d_e = 128$, Linear & 48M & 26.0  \\
    	$d_e = 256$, Linear & 53M & 27.1  \\	 
    	$d_e = 256$, 2-Layer Linear & 54M & 27.2 \\
    	$d_e = 128$, SAFE & 48M & 26.6  \\
    	$d_e = 256$, SAFE & 54M & 27.7  \\
    	\midrule[0.07em]
    	\citet{vaswani2017attention} & 65M & 27.3 \\
    	\textsc{Transformer-base} (reimpl.) & 61M & 27.7 \\
        \bottomrule[0.15em]
    \end{tabular}
    \caption{Experiments on the impact on SAFE vs a regular linear projection using \textsc{Transformer-base} on the WMT'14 EN-DE machine translation benchmark}
    \label{tb:safe}
\end{table}	
\begin{table}[t]
    \centering
    \footnotesize
    \begin{tabular}{lcc}
        \toprule [0.2em]
        \multirow{1}{*}{\textbf{\textsc{Model}}} %
        & \textbf{Param.} & \textbf{BLEU}
        \\ \midrule[0.1em]
        \textit{All-Shared} & 24M & 14.3  \\
        \textit{All-Shared, $d_m= 768$} & 41M & 22.0 \\
    	\textit{All-Shared (Independent FFN)} & 27M & 22.4  \\
    	\textit{All-Shared (except last)} & 31M & 23.2 \\
    	\textit{Every 2 layers shared} & 38M & 27.2 \\
    	\textsc{Sandwich} & 38M & 27.3 \\
    	\textsc{Sandwich, $L=8$} & 38M & \textbf{27.7} \\
    	\midrule[0.07em]
    	\citet{vaswani2017attention} & 65M & 27.3 \\
    	\textsc{Transformer-base} (Our reimpl.) & 61M & \textbf{27.7} \\
        \bottomrule[0.15em]
    \end{tabular}
    \caption{Experiments performed on WMT'14 EN-DE using different parameter sharing techniques.}
    \label{tb:prelim}
\end{table}
Given that the output of each Transformer layer depends directly on its two sub-layers ---multiheaded attention and the feedforward module--- when discussing alternatives for parameter sharing across transformer layers there are several options. As we aim to leverage the aforementioned properties of weight sharing, we performed preliminary experiments, investigating the capabilities of weight sharing in the following five settings: (1) \textbf{All-shared} Naively sharing all encoder and all decoder layers  ---that is including both of their sub-layers, following \citet{lan2020albert,dehghani2018universal}. (2) \textbf{All-Shared (Independent FFN)} Naively sharing all encoder and all decoder layers but allowing each layer $l \in [2, \cdot, L]$ to have an independent feed-forward sub-layer. (3) \textbf{All-Shared (except last)} Sharing weights across layers $l \in [1,\ldots, L-1]$ such that layer $L$ remains independent. (4) \textbf{Every 2 layers shared} Sharing every two layers, i.e. $[1,2],[3,4],[5,6]$ in the case of a 6-layer transformer. (5) \textbf{Sandwich} Finally, we only share the middle or central layers (i.e. $2 \leq l \leq L-1$), leaving layers $1$ and $L$ to have independent sets of parameters.

Table~\ref{tb:prelim} summarizes the results of our exploratory study. As can be seen, naive parameter sharing/tying approaches do not offer any advantages, hurting performance significantly ($\sim$50\%) when compared to the regular Transformer. However, our results also show that when combined properly, using Sandwich-style parameter sharing, we can attain a good balance of parameter reduction and performance. When compared to tasks such as pre-training deep contextualized word representations, generative tasks such as machine translation require informative token-level representations for each input token to be accurately translated. In this context, we surmise that the success of Sandwich-style parameter sharing on this sequence-to-sequence task is a result of it being able to have the input and output layers (arguably, the most important layers) be trained independently, allowing them to learn different operations than the sandwich layers. 
\paragraph{Combined Architecture} Having introduced our proposed parameter reduction techniques, we will now explain the Subformer architecture. The Subformer is composed of four main components, for both the encoder and decoder: the embedding layer, the model layers, the sandwich module and the projection layers. We disentangle the sandwiched layer dimension from that of the model layer, allowing the sandwich layer width to be larger than the rest of the model. For this reason, we denote the dimension of the sandwiched layer to be $d_s$ and its corresponding feed-forward dimension to be $\vec{d}_s$. %

\section{Experimental Setup}

We apply our method to a variety of sequence modeling tasks: neural machine translation, summarization, and language modeling. We compare  Transformer-base and big \citep{vaswani2017attention} with the Subformer trained in the same setting. We also include simple sandwich-style parameter sharing (denoted as Sandwich-\{base,big\}) and the usage of \textit{only} SAFE as an ablation of what these techniques do in their naive forms when decoupled. Additional implementation and training details with hyperparameter settings are in the appendix.

\paragraph{Machine Translation}
We evaluate our model on two standard translation benchmarks: WMT'14 English-German (EN-DE) and WMT'16 English-Romanian (EN-RO). Following previous work, we evaluate all models using tokenized BLEU \citep{papineni2002bleu}.  In order to better contextualize our results, we consider parameter-efficient models DeLighT \citep{mehta2020delight} (contemporaneous work), and the Evolved Transformer \citep{so2019evolved}.

\paragraph{Abstractive Summarization}
We test the model's ability to process long documents on the CNN-DailyMail summarization benchmark \citep{hermann2015teaching,nallapati2016abstractive}, comprising over 280K news articles paired with multi-sentence summaries. We also compare effects of BART \citep{lewis-etal-2020-bart} pretraining (details in Appendix~\ref{app:training_details}). For this task we contexutalize our results with specialized architectures such as Pointer-Generator Networks \citep{see2017get}, and methods leveraging pretraining: RobertaShare \citep{rothe-etal-2020-leveraging}, BertExtAbs \cite{liu2019text}, and BART \citep{lewis-etal-2020-bart}. We evaluate using ROUGE 1,2,L \citep{lin2004rouge}.

\paragraph{Language Modeling} 
We evaluate on the large-scale Wikitext-103 dataset \citep{merity2016pointer}. Models are evaluated in terms of perplexity on the test portion. In order to better contextualize our results, we consider the QRNN \citep{merity2018analysis}, Transformer-XL \citep{dai2019transformer} and Deep Equilibrium Model (DEQ) \citep{bai2019deep}, which also employs parameter sharing.

\section{Results}

\begin{table}[t]
    \centering
    \resizebox{0.49\textwidth}{!}{\begin{tabular}{lcccc}
        \toprule [0.2em]
        \multirow{2}{*}{\textbf{\textsc{Base Models}}} & \multicolumn{2}{c}{\bf WMT'14 EN-DE} & \multicolumn{2}{c}{\bf WMT'16 EN-RO}\\
        \cmidrule[0.15em](r){2-3}  \cmidrule[0.15em](r){4-5} 
    	& \textbf{Param.} & \textbf{BLEU} & \textbf{Param.} & \textbf{BLEU} \\ \midrule[0.1em]
    	DeLighT &  37M & 27.6 & 22M & 34.3\\
    	Evolved Transformer &  48M & 27.7 & --- & --- \\
    	DeLighT &  54M & 28.0 & 52M &\textbf{34.7} \\
    	Evolved Transformer&  64M & 28.2 & --- & --- \\
    	\midrule[0.07em]
    	\midrule[0.07em]
    	Transformer-base (orig) & 65M & 27.3& 62M & 34.2$^\dagger$\\
    	Transformer-base (ours) & 61M & 27.7 & 62M & 34.1 \\
        \midrule[0.1em]
    	Sandwich-base & 38M & 27.3 & --- & --- \\ 
    	Only SAFE, $d_e = 256$ & 54M & 27.7 & --- & --- \\
        \midrule[0.1em]
        \textsc{Subformer-small} & 38M & 27.7 & 20M & 34.1 \\
        \textsc{Subformer-base} & 52M & 28.1 & 48M & \textbf{34.7}\\
        \textsc{Subformer-mid} & 63M & \textbf{28.5} & --- & --- \\
        \bottomrule[0.15em]
    \end{tabular}}
    \caption{Results on WMT'14 EN-DE and WMT'16 EN-RO task, for our base models. The $^\dagger$ superscript indicates results from \citet{kasai2020nonautoregressive}.}
    \label{tb:mt-base}
\end{table} 
\begin{table}
    \centering 
    \footnotesize
    \begin{tabular}{lcc}
        \toprule [0.2em]
        \multirow{1}{*}{\textbf{\textsc{Big Models}}} %
        & \textbf{Param.} & \textbf{BLEU}\\
        \midrule[0.1em]
        Evolved Transformer &  222M & 29.0\\
        \midrule[0.07em]
        \midrule[0.07em]
        Transformer-big (orig) & 213M & 28.4  \\ 
        Transformer-big (ours) & 210M & 28.6 \\
        \midrule[0.1em]
    	Sandwich-big & 122M & 28.6 \\ 
        \midrule[0.1em]
        \textsc{Subformer-large} & 197M & \textbf{29.3} \\
        \bottomrule[0.15em]
    \end{tabular}
    \caption{Results on WMT'14 EN-DE for large models.} 
    \label{tb:mt-big}
\end{table} 
\paragraph{Machine Translation} Tables~\ref{tb:mt-base} and \ref{tb:mt-big}\footnote{In all tables, results from other work used to contextualize our results are placed above the double bar.} summarize our results on machine translation. Firstly, we note that our Transformer baselines outperform \citet{vaswani2017attention} (base: 27.3 $\rightarrow$ 27.7, big: 28.4 $\rightarrow$ 28.6). We surmise that this is due to training for longer and with a larger batch size. 

The Subformer outperforms our Transformer baselines when trained in the same setting, with similar or fewer parameters. \textsc{Subformer-small} reduces parameters by 40\%, matching performance with our Transformer baselines. \textsc{Subformer-base} and \textsc{mid}, outperform our model significantly (0.4, 0.8 BLEU) when using a fewer/similar number of parameters. Furthermore, we note that \textsc{Subformer-mid} performs similarly to the Transformer-big model (210M params.) in Table~\ref{tb:mt-big}, despite a 70\% parameter reduction.

For our large set of models (Table~\ref{tb:mt-big}), Sandwich-big achieves the same performance as our Transformer-big reimplementation, but with 40\% fewer parameters. We believe this to be an indication of the capability of Sandwich-style parameter sharing as the encoder/decoder layers get wider, while also providing further evidence for the overparameterization of the Transformer. Subformer-large, with 197M parameters achieves a significant 0.7 BLEU score gain over Transformer-big, despite using 13M fewer parameters. 

\paragraph{Language Modeling} Results for language modeling can be seen in Table~\ref{tb:langauge_modeling}. The Subformer uses adaptive input embeddings \citep{baevski2018adaptive} instead of SAFE embeddings, following common practice. We also train two Transformer baselines with the same setup ---one with the same amount of parameters and another with a similar parameter count to Transformer-XL--- to provide better context for comparison. Task-specific techniques that can be applied during training, such as caching \citep{dai2019transformer} or other methods applied during inference time \citep{Khandelwal2020Generalization,krause2018dynamic} can further improve all models so we do not focus on these.

\begin{table}[th]
    \centering
    \footnotesize
    \resizebox{0.45\textwidth}{!}{\begin{tabular}{lccc}
        \toprule [0.2em]
        \multirow{1}{*}{\textbf{\textsc{Model}}} %
    	& \textbf{Param.} & \textbf{CL} & \textbf{PPL} \\
    	\midrule[0.1em]
    	QRNN & 151M & --- & 33.00 \\
    	DeLighT & 99M & 480 & 24.14 \\
    	Transformer-XL & 151M & 640 & 24.03  \\
    	DEQ & 110M & --- & 23.20 \\
    	Adaptive Inputs & 247M & 480 & 19.03\\
    	\midrule[0.05em]
    	\midrule[0.05em]
    	Adaptive Inputs (4 Layer) & 96M & 480 & 26.42 \\
    	Adaptive Inputs (8 Layer) & 146M & 480 & 22.32 \\
    	\midrule[0.07em]
    	\textbf{\textsc{Subformer}} & \textbf{83M}& 480 & 20.88 \\
    	& 96M & 480 & 20.39 \\
    	& 122M & 480 & \textbf{19.90} \\
        \bottomrule[0.15em]
    \end{tabular}}
    \caption{Results on the Wikitext-103 benchmark. CL stands for Context Length.%
    } 
    \label{tb:langauge_modeling}
\end{table}	

As seen in Table~\ref{tb:langauge_modeling}, the Subformer outperforms the baselines by a significant margin (between 1.4 and 6.5 perplexity), with a significant reduction in parameters. 
\begin{table}[t]
    \centering
    \footnotesize
    \resizebox{0.49\textwidth}{!}{\begin{tabular}{lcccc}
        \toprule [0.2em]
    	\textbf{\textsc{Model}} & \textbf{Params.} & \textbf{R1} & \textbf{R2} & \textbf{RL}  
        \\ \midrule[0.1em]
        Ptr-Gen+Cov& --- &  39.5 & 17.3 & 36.4 \\
        RobertaShare & 152M & 40.3 & 18.9 & 37.6 \\
        BertExtAbs & 220M & 41.7 & 19.4 & 38.8 \\
        BART & 406M & 44.2 & 21.3 & 40.9 \\
        \midrule[0.05em]
        \midrule[0.05em]
        Transformer (3 Layer) & 57M & 40.0 & 17.5 & 36.7 \\
        Transformer & 77M & 40.1 & 17.6 & 36.8 \\
        Transformer-BART & 77M & 41.2 & 18.7 & 37.6 \\
        \midrule[0.1em]
        \textsc{Subformer-base} & 57M & 40.9 & 18.3 & 37.7 \\
        \textsc{Subformer-BART} & 57M & \textbf{41.6} & \textbf{19.2} & \textbf{38.4} \\
        \bottomrule[0.2em]
    \end{tabular}}
    \caption{Results on CNN-Daily Mail.} 
    \label{tb:sum}
\end{table}

\paragraph{Abstractive Summarization} For the CNN/Daily Mail summarization task we use Subformer-base. We also pretrain a Transformer and Subformer model with the same architecture on Wikipedia (details in Appendix~\ref{app:training_details}). As can be seen in Table~\ref{tb:sum}, the Subformer outperforms two Transformer baselines with both the same parameter count and its respective Transformer-base configuration in both settings, demonstrating the Subformer's performance on a variety of tasks and with longer sequences.

\paragraph{Discussion on Speed and Convergence} We found training time to consistently speed up by 10-30\%, and inference speed to either be faster by 10-20\% (keeping $d_s=d_m$) to be slightly slower by 10-30\% (when $d_s \gg d_m$) (due to more operations, similar to \citet{lan2020albert}). The Subformer converges faster most likely due to fewer parameters to optimize. Given the fewer number of parameters, it can be expected for the models to converge with fewer iterations. We test this on the task of language modeling, where we found that the Subformer converged 65\% faster than its Transformer counterpart, as shown in Table~\ref{tb:convergence}. We also measure inference speed on our machine translation models (Table~\ref{tb:inference}).
\begin{table}[t]
    \centering 
    \normalsize
    \scalebox{0.9}{\begin{tabular}{lccc}
        \toprule [0.2em]
        \multirow{1}{*}{Model} %
        & \textbf{Param.} & \textbf{Iterations} & {Dev. PPL}\\
        \midrule[0.1em]
        Adaptive Inputs & 146M & 272K & 22.31 \\ 
        \textsc{Subformer} & \textbf{83M} & \textbf{97K} & \textbf{20.84} \\
        \bottomrule[0.15em]
    \end{tabular}}
    \caption{Iterations to convergence on \textsc{Wikitext-103}}
    \label{tb:convergence}
\end{table} 
\begin{table}[t]
    \centering 
    \footnotesize
    \resizebox{!}{!}{\begin{tabular}{lccc}
        \toprule [0.2em]
        \multirow{1}{*}{Model} %
        & \textbf{Param.} & \textbf{Speed $\uparrow$} & BLEU\\
        \midrule[0.1em]
        DeLighT & 37M & 0.30x & 27.6 \\
        \midrule
        \midrule
        Transformer & 61M &  1.00x & 27.7 \\
        SAFE, $d_e=256$ & 54M & 1.17x & 27.7 \\
        \textsc{Sandwich-base} & 38M & 1.26x & 27.3 \\ 
        \textsc{Subformer-base} & 52M & 0.75x & 28.1 \\
        \bottomrule[0.15em]
    \end{tabular}}
    \caption{Inference speed for our models measured on a single V100 GPU on WMT'14 En-De (batch size: 384, 1.00x = 5135 tokens)}
    \label{tb:inference}
\end{table} 

\section{Conclusion}

In this paper we have presented the Subformer, a parameter-efficient Transformer-based generative model primarily based on two simple parameter factorization/sharing techniques. Our empirical results on three sequence modeling tasks show that the Subformer can achieve similar or better performance compared with a base/big Transformer with a $\sim$40\% parameter reduction. Furthermore, the simplicity of our approach
allows the Subformer to be used in conjunction with other parameter reduction techniques in the literature, for even smaller but performant models. We hope our work incites interest in parameter sharing techniques for an even wider range of Transformer models, ultimately helping reduce their computational cost in general.

\section*{Ethical Considerations}
This work has impact in the field of natural language processing, and develops a more efficient approach for learning performant generative models. As with much of language technology has the potential to be both used for good and used maliciously. We also experiment with pretraining, learning representations in an unsupervised way, which is likely to capture and amplify biases found in the
data. However, our approach has a potential positive impact given the lower cost and energy expenditure needed to train our proposed model.
\section*{Acknowledgments}
We thank Jorge Balazs, Yusuke Iwasawa, Jungo Kasai, Cristian Rodriguez-Opazo, Alfredo Solano, Yutaro Yamada, and Victor Zhong for their helpful feedback and discussions over this work. MR is grateful to the Masason Foundation for their support.

\bibliographystyle{acl_natbib}
\bibliography{acl2021,anthology}

\begin{thebibliography}{44}
\expandafter\ifx\csname natexlab\endcsname\relax\def\natexlab#1{#1}\fi

\bibitem[{Baevski and Auli(2019)}]{baevski2018adaptive}
Alexei Baevski and Michael Auli. 2019.
\newblock \href {https://openreview.net/forum?id=ByxZX20qFQ} {Adaptive input
  representations for neural language modeling}.
\newblock In \emph{7th International Conference on Learning Representations,
  {ICLR} 2019, New Orleans, LA, USA, May 6-9, 2019}. OpenReview.net.

\bibitem[{Bai et~al.(2019)Bai, Kolter, and Koltun}]{bai2019deep}
Shaojie Bai, J.~Zico Kolter, and Vladlen Koltun. 2019.
\newblock \href
  {https://proceedings.neurips.cc/paper/2019/hash/01386bd6d8e091c2ab4c7c7de644d37b-Abstract.html}
  {Deep equilibrium models}.
\newblock In \emph{Advances in Neural Information Processing Systems 32: Annual
  Conference on Neural Information Processing Systems 2019, NeurIPS 2019,
  December 8-14, 2019, Vancouver, BC, Canada}, pages 688--699.

\bibitem[{Cheong and Daniel(2019)}]{cheong2019transformers}
Robin Cheong and Robel Daniel. 2019.
\newblock Transformers. zip: {{Compressing}} transformers with pruning and
  quantization.
\newblock Technical report, {Stanford University, Stanford, California}.

\bibitem[{Dai et~al.(2019)Dai, Yang, Yang, Carbonell, Le, and
  Salakhutdinov}]{dai2019transformer}
Zihang Dai, Zhilin Yang, Yiming Yang, Jaime Carbonell, Quoc Le, and Ruslan
  Salakhutdinov. 2019.
\newblock \href {https://doi.org/10.18653/v1/P19-1285} {Transformer-{XL}:
  Attentive language models beyond a fixed-length context}.
\newblock In \emph{Proceedings of the 57th Annual Meeting of the Association
  for Computational Linguistics}, pages 2978--2988, Florence, Italy.
  Association for Computational Linguistics.

\bibitem[{Dehghani et~al.(2019)Dehghani, Gouws, Vinyals, Uszkoreit, and
  Kaiser}]{dehghani2018universal}
Mostafa Dehghani, Stephan Gouws, Oriol Vinyals, Jakob Uszkoreit, and Lukasz
  Kaiser. 2019.
\newblock \href {https://openreview.net/forum?id=HyzdRiR9Y7} {Universal
  transformers}.
\newblock In \emph{7th International Conference on Learning Representations,
  {ICLR} 2019, New Orleans, LA, USA, May 6-9, 2019}. OpenReview.net.

\bibitem[{Devlin et~al.(2019)Devlin, Chang, Lee, and
  Toutanova}]{devlin2019bert}
Jacob Devlin, Ming-Wei Chang, Kenton Lee, and Kristina Toutanova. 2019.
\newblock \href {https://doi.org/10.18653/v1/N19-1423} {{BERT}: Pre-training of
  deep bidirectional transformers for language understanding}.
\newblock In \emph{Proceedings of the 2019 Conference of the North {A}merican
  Chapter of the Association for Computational Linguistics: Human Language
  Technologies, Volume 1 (Long and Short Papers)}, pages 4171--4186,
  Minneapolis, Minnesota. Association for Computational Linguistics.

\bibitem[{Dou et~al.(2018)Dou, Tu, Wang, Shi, and Zhang}]{Dou_2018}
Zi-Yi Dou, Zhaopeng Tu, Xing Wang, Shuming Shi, and Tong Zhang. 2018.
\newblock \href {https://doi.org/10.18653/v1/D18-1457} {Exploiting deep
  representations for neural machine translation}.
\newblock In \emph{Proceedings of the 2018 Conference on Empirical Methods in
  Natural Language Processing}, pages 4253--4262, Brussels, Belgium.
  Association for Computational Linguistics.

\bibitem[{Edunov et~al.(2019)Edunov, Baevski, and Auli}]{edunov2019pre}
Sergey Edunov, Alexei Baevski, and Michael Auli. 2019.
\newblock \href {https://doi.org/10.18653/v1/N19-1409} {Pre-trained language
  model representations for language generation}.
\newblock In \emph{Proceedings of the 2019 Conference of the North {A}merican
  Chapter of the Association for Computational Linguistics: Human Language
  Technologies, Volume 1 (Long and Short Papers)}, pages 4052--4059,
  Minneapolis, Minnesota. Association for Computational Linguistics.

\bibitem[{Fan et~al.(2020)Fan, Grave, and Joulin}]{fan2020reducing}
Angela Fan, Edouard Grave, and Armand Joulin. 2020.
\newblock \href {https://openreview.net/forum?id=SylO2yStDr} {Reducing
  transformer depth on demand with structured dropout}.
\newblock In \emph{8th International Conference on Learning Representations,
  {ICLR} 2020, Addis Ababa, Ethiopia, April 26-30, 2020}. OpenReview.net.

\bibitem[{Ganesh et~al.(2020)Ganesh, Chen, Lou, Khan, Yang, Chen, Winslett,
  Sajjad, and Nakov}]{ganeshCompressingLargeScaleTransformerBased2020}
Prakhar Ganesh, Yao Chen, Xin Lou, Mohammad~Ali Khan, Yin Yang, Deming Chen,
  Marianne Winslett, Hassan Sajjad, and Preslav Nakov. 2020.
\newblock \href {http://arxiv.org/abs/2002.11985} {Compressing
  {{Large}}-{{Scale Transformer}}-{{Based Models}}: {{A Case Study}} on
  {{BERT}}}.
\newblock \emph{arXiv:2002.11985 [cs, stat]}.

\bibitem[{Ghazvininejad et~al.(2019)Ghazvininejad, Levy, Liu, and
  Zettlemoyer}]{ghazvininejad2019mask}
Marjan Ghazvininejad, Omer Levy, Yinhan Liu, and Luke Zettlemoyer. 2019.
\newblock \href {https://doi.org/10.18653/v1/D19-1633} {Mask-predict: Parallel
  decoding of conditional masked language models}.
\newblock In \emph{Proceedings of the 2019 Conference on Empirical Methods in
  Natural Language Processing and the 9th International Joint Conference on
  Natural Language Processing (EMNLP-IJCNLP)}, pages 6112--6121, Hong Kong,
  China. Association for Computational Linguistics.

\bibitem[{Hermann et~al.(2015)Hermann, Kocisk{\'{y}}, Grefenstette, Espeholt,
  Kay, Suleyman, and Blunsom}]{hermann2015teaching}
Karl~Moritz Hermann, Tom{\'{a}}s Kocisk{\'{y}}, Edward Grefenstette, Lasse
  Espeholt, Will Kay, Mustafa Suleyman, and Phil Blunsom. 2015.
\newblock \href
  {https://proceedings.neurips.cc/paper/2015/hash/afdec7005cc9f14302cd0474fd0f3c96-Abstract.html}
  {Teaching machines to read and comprehend}.
\newblock In \emph{Advances in Neural Information Processing Systems 28: Annual
  Conference on Neural Information Processing Systems 2015, December 7-12,
  2015, Montreal, Quebec, Canada}, pages 1693--1701.

\bibitem[{Kasai et~al.(2020)Kasai, Cross, Ghazvininejad, and
  Gu}]{kasai2020nonautoregressive}
Jungo Kasai, James Cross, Marjan Ghazvininejad, and Jiatao Gu. 2020.
\newblock \href {http://proceedings.mlr.press/v119/kasai20a.html}
  {Non-autoregressive machine translation with disentangled context
  transformer}.
\newblock In \emph{Proceedings of the 37th International Conference on Machine
  Learning, {ICML} 2020, 13-18 July 2020, Virtual Event}, volume 119 of
  \emph{Proceedings of Machine Learning Research}, pages 5144--5155. {PMLR}.

\bibitem[{Khandelwal et~al.(2020)Khandelwal, Levy, Jurafsky, Zettlemoyer, and
  Lewis}]{Khandelwal2020Generalization}
Urvashi Khandelwal, Omer Levy, Dan Jurafsky, Luke Zettlemoyer, and Mike Lewis.
  2020.
\newblock \href {https://openreview.net/forum?id=HklBjCEKvH} {Generalization
  through memorization: Nearest neighbor language models}.
\newblock In \emph{8th International Conference on Learning Representations,
  {ICLR} 2020, Addis Ababa, Ethiopia, April 26-30, 2020}. OpenReview.net.

\bibitem[{Kingma and Ba(2015)}]{kingma2014adam}
Diederik~P. Kingma and Jimmy Ba. 2015.
\newblock \href {http://arxiv.org/abs/1412.6980} {Adam: {A} method for
  stochastic optimization}.
\newblock In \emph{3rd International Conference on Learning Representations,
  {ICLR} 2015, San Diego, CA, USA, May 7-9, 2015, Conference Track
  Proceedings}.

\bibitem[{Kitaev et~al.(2020)Kitaev, Kaiser, and Levskaya}]{kitaev2020reformer}
Nikita Kitaev, Lukasz Kaiser, and Anselm Levskaya. 2020.
\newblock \href {https://openreview.net/forum?id=rkgNKkHtvB} {Reformer: The
  efficient transformer}.
\newblock In \emph{8th International Conference on Learning Representations,
  {ICLR} 2020, Addis Ababa, Ethiopia, April 26-30, 2020}. OpenReview.net.

\bibitem[{Krause et~al.(2018)Krause, Kahembwe, Murray, and
  Renals}]{krause2018dynamic}
Ben Krause, Emmanuel Kahembwe, Iain Murray, and Steve Renals. 2018.
\newblock \href {http://proceedings.mlr.press/v80/krause18a.html} {Dynamic
  evaluation of neural sequence models}.
\newblock In \emph{Proceedings of the 35th International Conference on Machine
  Learning}, volume~80 of \emph{Proceedings of Machine Learning Research},
  pages 2766--2775, Stockholmsmässan, Stockholm Sweden. PMLR.

\bibitem[{Lan et~al.(2020)Lan, Chen, Goodman, Gimpel, Sharma, and
  Soricut}]{lan2020albert}
Zhenzhong Lan, Mingda Chen, Sebastian Goodman, Kevin Gimpel, Piyush Sharma, and
  Radu Soricut. 2020.
\newblock \href {https://openreview.net/forum?id=H1eA7AEtvS} {{ALBERT:} {A}
  lite {BERT} for self-supervised learning of language representations}.
\newblock In \emph{8th International Conference on Learning Representations,
  {ICLR} 2020, Addis Ababa, Ethiopia, April 26-30, 2020}. OpenReview.net.

\bibitem[{Lee et~al.(2018)Lee, Mansimov, and Cho}]{lee2018deterministic}
Jason Lee, Elman Mansimov, and Kyunghyun Cho. 2018.
\newblock \href {https://doi.org/10.18653/v1/D18-1149} {Deterministic
  non-autoregressive neural sequence modeling by iterative refinement}.
\newblock In \emph{Proceedings of the 2018 Conference on Empirical Methods in
  Natural Language Processing}, pages 1173--1182, Brussels, Belgium.
  Association for Computational Linguistics.

\bibitem[{Lee et~al.(2019)Lee, Ajanthan, and
  Torr}]{leeSNIPSingleshotNetwork2018}
Namhoon Lee, Thalaiyasingam Ajanthan, and Philip H.~S. Torr. 2019.
\newblock \href {https://openreview.net/forum?id=B1VZqjAcYX} {Snip: single-shot
  network pruning based on connection sensitivity}.
\newblock In \emph{7th International Conference on Learning Representations,
  {ICLR} 2019, New Orleans, LA, USA, May 6-9, 2019}. OpenReview.net.

\bibitem[{Lewis et~al.(2020)Lewis, Liu, Goyal, Ghazvininejad, Mohamed, Levy,
  Stoyanov, and Zettlemoyer}]{lewis-etal-2020-bart}
Mike Lewis, Yinhan Liu, Naman Goyal, Marjan Ghazvininejad, Abdelrahman Mohamed,
  Omer Levy, Veselin Stoyanov, and Luke Zettlemoyer. 2020.
\newblock \href {https://doi.org/10.18653/v1/2020.acl-main.703} {{BART}:
  Denoising sequence-to-sequence pre-training for natural language generation,
  translation, and comprehension}.
\newblock In \emph{Proceedings of the 58th Annual Meeting of the Association
  for Computational Linguistics}, pages 7871--7880, Online. Association for
  Computational Linguistics.

\bibitem[{Lin(2004)}]{lin2004rouge}
Chin-Yew Lin. 2004.
\newblock \href {https://www.aclweb.org/anthology/W04-1013} {{ROUGE}: A package
  for automatic evaluation of summaries}.
\newblock In \emph{Text Summarization Branches Out}, pages 74--81, Barcelona,
  Spain. Association for Computational Linguistics.

\bibitem[{Liu and Lapata(2019)}]{liu2019text}
Yang Liu and Mirella Lapata. 2019.
\newblock Text summarization with pretrained encoders.
\newblock \emph{arXiv preprint arXiv:1908.08345}.

\bibitem[{Mehta et~al.(2020{\natexlab{a}})Mehta, Ghazvininejad, Iyer,
  Zettlemoyer, and Hajishirzi}]{mehta2020delight}
Sachin Mehta, Marjan Ghazvininejad, Srinivasan Iyer, Luke Zettlemoyer, and
  Hannaneh Hajishirzi. 2020{\natexlab{a}}.
\newblock \href {http://arxiv.org/abs/2008.00623} {Delight: Very deep and
  light-weight transformer}.

\bibitem[{Mehta et~al.(2020{\natexlab{b}})Mehta, Koncel{-}Kedziorski,
  Rastegari, and Hajishirzi}]{mehta2020define}
Sachin Mehta, Rik Koncel{-}Kedziorski, Mohammad Rastegari, and Hannaneh
  Hajishirzi. 2020{\natexlab{b}}.
\newblock \href {https://openreview.net/forum?id=rJeXS04FPH} {Define: Deep
  factorized input token embeddings for neural sequence modeling}.
\newblock In \emph{8th International Conference on Learning Representations,
  {ICLR} 2020, Addis Ababa, Ethiopia, April 26-30, 2020}. OpenReview.net.

\bibitem[{Merity et~al.(2018)Merity, Keskar, and Socher}]{merity2018analysis}
Stephen Merity, Nitish~Shirish Keskar, and Richard Socher. 2018.
\newblock \href {http://arxiv.org/abs/1803.08240} {An analysis of neural
  language modeling at multiple scales}.

\bibitem[{Merity et~al.(2017)Merity, Xiong, Bradbury, and
  Socher}]{merity2016pointer}
Stephen Merity, Caiming Xiong, James Bradbury, and Richard Socher. 2017.
\newblock \href {https://openreview.net/forum?id=Byj72udxe} {Pointer sentinel
  mixture models}.
\newblock In \emph{5th International Conference on Learning Representations,
  {ICLR} 2017, Toulon, France, April 24-26, 2017, Conference Track
  Proceedings}. OpenReview.net.

\bibitem[{Micikevicius et~al.(2018)Micikevicius, Narang, Alben, Diamos, Elsen,
  Garc{\'{\i}}a, Ginsburg, Houston, Kuchaiev, Venkatesh, and
  Wu}]{micikevicius2017mixed}
Paulius Micikevicius, Sharan Narang, Jonah Alben, Gregory~F. Diamos, Erich
  Elsen, David Garc{\'{\i}}a, Boris Ginsburg, Michael Houston, Oleksii
  Kuchaiev, Ganesh Venkatesh, and Hao Wu. 2018.
\newblock \href {https://openreview.net/forum?id=r1gs9JgRZ} {Mixed precision
  training}.
\newblock In \emph{6th International Conference on Learning Representations,
  {ICLR} 2018, Vancouver, BC, Canada, April 30 - May 3, 2018, Conference Track
  Proceedings}. OpenReview.net.

\bibitem[{Nallapati et~al.(2016)Nallapati, Zhou, dos Santos, Gul{\c{c}}ehre,
  and Xiang}]{nallapati2016abstractive}
Ramesh Nallapati, Bowen Zhou, Cicero dos Santos, {\c{C}}a{\u{g}}lar
  Gul{\c{c}}ehre, and Bing Xiang. 2016.
\newblock \href {https://doi.org/10.18653/v1/K16-1028} {Abstractive text
  summarization using sequence-to-sequence {RNN}s and beyond}.
\newblock In \emph{Proceedings of The 20th {SIGNLL} Conference on Computational
  Natural Language Learning}, pages 280--290, Berlin, Germany. Association for
  Computational Linguistics.

\bibitem[{Ott et~al.(2019)Ott, Edunov, Baevski, Fan, Gross, Ng, Grangier, and
  Auli}]{ott2019fairseq}
Myle Ott, Sergey Edunov, Alexei Baevski, Angela Fan, Sam Gross, Nathan Ng,
  David Grangier, and Michael Auli. 2019.
\newblock \href {https://doi.org/10.18653/v1/N19-4009} {fairseq: A fast,
  extensible toolkit for sequence modeling}.
\newblock In \emph{Proceedings of the 2019 Conference of the North {A}merican
  Chapter of the Association for Computational Linguistics (Demonstrations)},
  pages 48--53, Minneapolis, Minnesota. Association for Computational
  Linguistics.

\bibitem[{Papineni et~al.(2002)Papineni, Roukos, Ward, and
  Zhu}]{papineni2002bleu}
Kishore Papineni, Salim Roukos, Todd Ward, and Wei-Jing Zhu. 2002.
\newblock \href {https://doi.org/10.3115/1073083.1073135} {{B}leu: a method for
  automatic evaluation of machine translation}.
\newblock In \emph{Proceedings of the 40th Annual Meeting of the Association
  for Computational Linguistics}, pages 311--318, Philadelphia, Pennsylvania,
  USA. Association for Computational Linguistics.

\bibitem[{Paszke et~al.(2019)Paszke, Gross, Massa, Lerer, Bradbury, Chanan,
  Killeen, Lin, Gimelshein, Antiga, Desmaison, K{\"{o}}pf, Yang, DeVito,
  Raison, Tejani, Chilamkurthy, Steiner, Fang, Bai, and
  Chintala}]{paszke_pytorch_2019}
Adam Paszke, Sam Gross, Francisco Massa, Adam Lerer, James Bradbury, Gregory
  Chanan, Trevor Killeen, Zeming Lin, Natalia Gimelshein, Luca Antiga, Alban
  Desmaison, Andreas K{\"{o}}pf, Edward Yang, Zachary DeVito, Martin Raison,
  Alykhan Tejani, Sasank Chilamkurthy, Benoit Steiner, Lu~Fang, Junjie Bai, and
  Soumith Chintala. 2019.
\newblock \href
  {https://proceedings.neurips.cc/paper/2019/hash/bdbca288fee7f92f2bfa9f7012727740-Abstract.html}
  {Pytorch: An imperative style, high-performance deep learning library}.
\newblock In \emph{Advances in Neural Information Processing Systems 32: Annual
  Conference on Neural Information Processing Systems 2019, NeurIPS 2019,
  December 8-14, 2019, Vancouver, BC, Canada}, pages 8024--8035.

\bibitem[{Peters et~al.(2018)Peters, Neumann, Iyyer, Gardner, Clark, Lee, and
  Zettlemoyer}]{Peters:2018}
Matthew Peters, Mark Neumann, Mohit Iyyer, Matt Gardner, Christopher Clark,
  Kenton Lee, and Luke Zettlemoyer. 2018.
\newblock \href {https://doi.org/10.18653/v1/N18-1202} {Deep contextualized
  word representations}.
\newblock In \emph{Proceedings of the 2018 Conference of the North {A}merican
  Chapter of the Association for Computational Linguistics: Human Language
  Technologies, Volume 1 (Long Papers)}, pages 2227--2237, New Orleans,
  Louisiana. Association for Computational Linguistics.

\bibitem[{Prato et~al.(2019)Prato, Charlaix, and
  Rezagholizadeh}]{pratoFullyQuantizedTransformer2019}
Gabriele Prato, Ella Charlaix, and M.~Rezagholizadeh. 2019.
\newblock Fully {{Quantized Transformer}} for {{Improved Translation}}.
\newblock \emph{ArXiv}.

\bibitem[{Press and Wolf(2017)}]{press-wolf-2017-using}
Ofir Press and Lior Wolf. 2017.
\newblock \href {https://www.aclweb.org/anthology/E17-2025} {Using the output
  embedding to improve language models}.
\newblock In \emph{Proceedings of the 15th Conference of the {E}uropean Chapter
  of the Association for Computational Linguistics: Volume 2, Short Papers},
  pages 157--163, Valencia, Spain. Association for Computational Linguistics.

\bibitem[{Rothe et~al.(2020)Rothe, Narayan, and
  Severyn}]{rothe-etal-2020-leveraging}
Sascha Rothe, Shashi Narayan, and Aliaksei Severyn. 2020.
\newblock \href {https://doi.org/10.1162/tacl_a_00313} {Leveraging pre-trained
  checkpoints for sequence generation tasks}.
\newblock \emph{Transactions of the Association for Computational Linguistics},
  8:264--280.

\bibitem[{See et~al.(2017)See, Liu, and Manning}]{see2017get}
Abigail See, Peter~J. Liu, and Christopher~D. Manning. 2017.
\newblock \href {https://doi.org/10.18653/v1/P17-1099} {Get to the point:
  Summarization with pointer-generator networks}.
\newblock In \emph{Proceedings of the 55th Annual Meeting of the Association
  for Computational Linguistics (Volume 1: Long Papers)}, pages 1073--1083,
  Vancouver, Canada. Association for Computational Linguistics.

\bibitem[{Sennrich et~al.(2016)Sennrich, Haddow, and
  Birch}]{sennrich2016neural}
Rico Sennrich, Barry Haddow, and Alexandra Birch. 2016.
\newblock \href {https://doi.org/10.18653/v1/P16-1162} {Neural machine
  translation of rare words with subword units}.
\newblock In \emph{Proceedings of the 54th Annual Meeting of the Association
  for Computational Linguistics (Volume 1: Long Papers)}, pages 1715--1725,
  Berlin, Germany. Association for Computational Linguistics.

\bibitem[{So et~al.(2019)So, Le, and Liang}]{so2019evolved}
David~R. So, Quoc~V. Le, and Chen Liang. 2019.
\newblock \href {http://proceedings.mlr.press/v97/so19a.html} {The evolved
  transformer}.
\newblock In \emph{Proceedings of the 36th International Conference on Machine
  Learning, {ICML} 2019, 9-15 June 2019, Long Beach, California, {USA}},
  volume~97 of \emph{Proceedings of Machine Learning Research}, pages
  5877--5886. {PMLR}.

\bibitem[{Sutskever et~al.(2013)Sutskever, Martens, Dahl, and
  Hinton}]{sutskever2013importance}
Ilya Sutskever, James Martens, George~E. Dahl, and Geoffrey~E. Hinton. 2013.
\newblock \href {http://proceedings.mlr.press/v28/sutskever13.html} {On the
  importance of initialization and momentum in deep learning}.
\newblock In \emph{Proceedings of the 30th International Conference on Machine
  Learning, {ICML} 2013, Atlanta, GA, USA, 16-21 June 2013}, volume~28 of
  \emph{{JMLR} Workshop and Conference Proceedings}, pages 1139--1147.
  JMLR.org.

\bibitem[{Vaswani et~al.(2017)Vaswani, Shazeer, Parmar, Uszkoreit, Jones,
  Gomez, Kaiser, and Polosukhin}]{vaswani2017attention}
Ashish Vaswani, Noam Shazeer, Niki Parmar, Jakob Uszkoreit, Llion Jones,
  Aidan~N. Gomez, Lukasz Kaiser, and Illia Polosukhin. 2017.
\newblock \href
  {https://proceedings.neurips.cc/paper/2017/hash/3f5ee243547dee91fbd053c1c4a845aa-Abstract.html}
  {Attention is all you need}.
\newblock In \emph{Advances in Neural Information Processing Systems 30: Annual
  Conference on Neural Information Processing Systems 2017, December 4-9, 2017,
  Long Beach, CA, {USA}}, pages 5998--6008.

\bibitem[{Wu et~al.(2019)Wu, Fan, Baevski, Dauphin, and Auli}]{wu2019pay}
Felix Wu, Angela Fan, Alexei Baevski, Yann~N. Dauphin, and Michael Auli. 2019.
\newblock \href {https://openreview.net/forum?id=SkVhlh09tX} {Pay less
  attention with lightweight and dynamic convolutions}.
\newblock In \emph{7th International Conference on Learning Representations,
  {ICLR} 2019, New Orleans, LA, USA, May 6-9, 2019}. OpenReview.net.

\bibitem[{Wu et~al.(2020)Wu, Liu, Lin, Lin, and Han}]{wu2020lite}
Zhanghao Wu, Zhijian Liu, Ji~Lin, Yujun Lin, and Song Han. 2020.
\newblock \href {https://openreview.net/forum?id=ByeMPlHKPH} {Lite transformer
  with long-short range attention}.
\newblock In \emph{8th International Conference on Learning Representations,
  {ICLR} 2020, Addis Ababa, Ethiopia, April 26-30, 2020}. OpenReview.net.

\bibitem[{Zafrir et~al.(2019)Zafrir, Boudoukh, Izsak, and
  Wasserblat}]{zafrirQ8BERTQuantized8Bit2019}
Ofir Zafrir, Guy Boudoukh, Peter Izsak, and Moshe Wasserblat. 2019.
\newblock \href {http://arxiv.org/abs/1910.06188} {{{Q8BERT}}: {{Quantized 8Bit
  BERT}}}.
\newblock \emph{arXiv:1910.06188 [cs]}.

\end{thebibliography}
\newpage~\newpage
\insertappendix
\end{document}